\documentclass[conference]{IEEEtran}
\IEEEoverridecommandlockouts
\usepackage{cite}
\usepackage{amsmath,amssymb,amsfonts}
\usepackage{algorithmic}
\usepackage{graphicx}
\usepackage{textcomp}
\usepackage{xcolor}
\usepackage{booktabs}
\usepackage{float}
\usepackage{balance}
\usepackage{multirow}
\usepackage[ruled,vlined,linesnumbered]{algorithm2e}

\begin{document}

\title{FELA: A Multi-Agent Evolutionary System for Feature Engineering of Industrial Event Log Data}

\author{\IEEEauthorblockN{Kun Ouyang*}
\IEEEauthorblockA{\textit{LIGHTSPEED STUDIOS,} \\
\thanks{*Equal contribution. \textsuperscript{\textdagger} corresponding author.}
Singapore \\
ouyangkun@u.nus.edu}
\and
\IEEEauthorblockN{Haoyu Wang*}
\IEEEauthorblockA{\textit{Department of Electronic Engineering,} \\
\textit{Tsinghua University}\\
\textit{LIGHTSPEED STUDIOS,} \\ 
China \\
wanghy22@mails.tsinghua.edu.cn;}
\and
\IEEEauthorblockN{Dong Fang\textsuperscript{\textdagger}}
\IEEEauthorblockA{\textit{LIGHTSPEED STUDIOS,} \\
Shenzhen, China \\
df572@outlook.com}
}

\maketitle

\begin{abstract}
Event log data, recording fine-grained user actions and system events, represent one of the most valuable assets for modern digital services. However, the complexity and heterogeneity of industrial event logs—characterized by large scale, high dimensionality, diverse data types, and intricate temporal or relational structures—make feature engineering extremely challenging. Existing automatic feature engineering approaches, such as AutoML or genetic methods, often suffer from limited explainability, rigid predefined operations, and poor adaptability to complicated heterogeneous data.

In this paper, we propose FELA (Feature Engineering LLM Agents), a multi-agent evolutionary system that autonomously extracts meaningful and high-performing features from complex industrial event log data. FELA integrates the reasoning and coding capabilities of large language models (LLMs) with an insight-guided self-evolution paradigm. Specifically, FELA employs specialized agents—Idea Agents, Code Agents, and Critic Agents—to collaboratively generate, validate, and implement novel feature ideas. An Evaluation Agent summarizes feedback and updates a hierarchical knowledge base and dual-memory system to enable continual improvement. Moreover, FELA introduces an agentic evolution algorithm, combining reinforcement learning and genetic algorithm principles to balance exploration and exploitation across the idea space.

Extensive experiments on real industrial datasets demonstrate that FELA can generate explainable, domain-relevant features that significantly improve model performance while reducing manual effort. Our results highlight the potential of LLM-based multi-agent systems as a general framework for automated, interpretable, and adaptive feature engineering in complex real-world environments.
\end{abstract}

\begin{IEEEkeywords}
feature engineering, agentic learning, LLM
\end{IEEEkeywords}

\section{Introduction}
Event logs have become one of the most valuable assets of modern enterprises, capturing rich traces of user behavior and system operations across large-scale services. These logs are massive—often spanning terabytes per day—and exhibit highly heterogeneous and temporal structures. Unlike standard tabular data, where each record is typically treated as an independent and identically distributed (i.i.d.) instance, event logs encode sequential dependencies and contextual relationships across time. This intrinsic complexity makes them substantially more challenging to analyze, yet also more informative. Each record may involve multiple dimensions with distinct semantics, cardinalities, and data types. For instance, in mobile gaming, a player’s log may contain timestamps, categorical attributes, integer-valued statistics (e.g., kills), and continuous metrics (e.g., kill–death ratios). In contrast, e-commerce platforms record user click histories, campaign identifiers, and detailed product attributes. Such diversity and scale make event logs both highly informative and extremely challenging to analyze.

Extracting meaningful knowledge from these logs relies heavily on feature engineering, where data scientists manually design informative variables guided by domain expertise and business understanding~\cite{KaggleSurvey2021}. This process involves extensive data exploration, reasoning about domain logic, hypothesis formulation, and iterative experimentation—often consuming days or weeks of effort even for experienced practitioners. The challenge becomes even more severe in industrial environments serving millions of users, where the sheer data volume and heterogeneity amplify the cost and complexity. As a result, there is a long-standing demand for technologies that can automate this process, leading to the field of \textit{Automated Feature Engineering} (AutoFE).

Existing AutoFE research spans diverse directions, including AutoML~\cite{liu2018progressive}, genetic algorithms (GA)~\cite{qi2023auto,real2019regularized}, and reinforcement learning (RL)~\cite{khurana2018feature,williams1992simple}. These methods aim to replace the manual trial-and-error cycle with self-improvement mechanisms. However, their lack of explainability and generality has significantly hindered industrial adoption. Recently, the emergence of large language model (LLM)–based agents has opened a promising new direction that unifies automation with human-like reasoning and interpretability. LLMs possess broad prior knowledge across domains, strong coding abilities comparable to human engineers, and the capacity to perform structured reasoning within complex contexts. They have demonstrated potential in data-centric tasks such as scientific discovery~\cite{ai-researcher,alpha-evolve}, mathematical problem solving~\cite{mm-agent}, and symbolic regression~\cite{llm-srbench}. Nonetheless, current applications of LLMs to feature engineering remain limited to simple tabular data~\cite{dynamicllm,llmfe}, leaving the challenge of building robust and general systems for real-world industrial scenarios largely unresolved.

To bridge this gap, we present a multi-agent collaborative system that automatically extracts novel and insightful features from large-scale industrial event log data. Our system is designed to address the following key challenges:

\begin{itemize}
    \item \textbf{Data complexity.} Event log data exhibit rich temporal dependencies and heterogeneous structures that differ fundamentally from standard tabular data. Many logs also encode implicit relational patterns. For example, in mobile games, players sharing the same \texttt{TeamID} form a teammate network critical for retention analysis; in social applications, comment logs form dynamic interaction graphs; and in e-commerce platforms, user–item activities naturally form bipartite graphs. A robust solution must therefore exploit both temporal and relational structure when constructing features. Furthermore, real-world event logs often include attributes stored in diverse formats (e.g., lists, JSON strings, or embedded URLs), requiring extensive preprocessing before feature extraction. Even expert data scientists spend substantial effort resolving such heterogeneity, and transferring expertise across domains (e.g., from gaming to e-commerce) demands considerable relearning. These factors collectively constrain the scalability and generalizability of traditional automatic feature engineering methods.

    \item \textbf{Explainability.} Beyond managing complexity, ensuring feature explainability is essential for both scientific rigor and industrial deployment. Diverse stakeholders—including data engineers, domain experts, and decision makers—must be able to interpret the semantics and rationale of generated features. This requires not only feature-level transparency (i.e., understanding what a feature represents) but also process-level transparency (i.e., how it was derived). In practice, the most valuable features often capture deep relationships between data structure, semantics, and business logic. Without a clear mechanism to encode and communicate such reasoning, automated feature engineering risks producing trivial or misleading results. Moreover, given the structural and semantic complexity of industrial logs, explainability becomes crucial for tracing, validating, and refining the feature derivation process.

    \item \textbf{Self-evolution.} The high dimensionality and combinatorial nature of event log data create an enormous feature search space, which poses a fundamental challenge for traditional automated methods. Classical approaches typically rely on predefined transformation operators and heuristic search strategies, yet the vast combination space makes it difficult to identify promising exploration directions. Reinforcement- or evolution-based frameworks further suffer from sparse or delayed reward signals relative to the size of the search space, resulting in inefficient exploration and suboptimal feature discovery. Addressing this challenge requires a mechanism that integrates evaluation feedback with knowledge distilled from historical trajectories, allowing the search process to leverage past experience and avoid inefficient wandering in the immense feature space.
\end{itemize}

To address the aforementioned challenges, we propose a comprehensive multi-agent collaborative system for automatic feature extraction from industrial event log data, termed \textbf{Feature Engineering LLM Agents (FELA)}. FELA autonomously discovers novel and high-performing features through an \textit{insight-guided self-evolution} paradigm. Beyond constructing an agent workflow that merely generates feature engineering code, FELA tackles a more fundamental question: \textit{How can ideas be systematically and controllably evolved with prior knowledge?} 

The system is designed around three key components that collectively enable effective feature discovery on complex, large-scale datasets.

\begin{figure}[t]
  \includegraphics[width=0.53\textwidth]{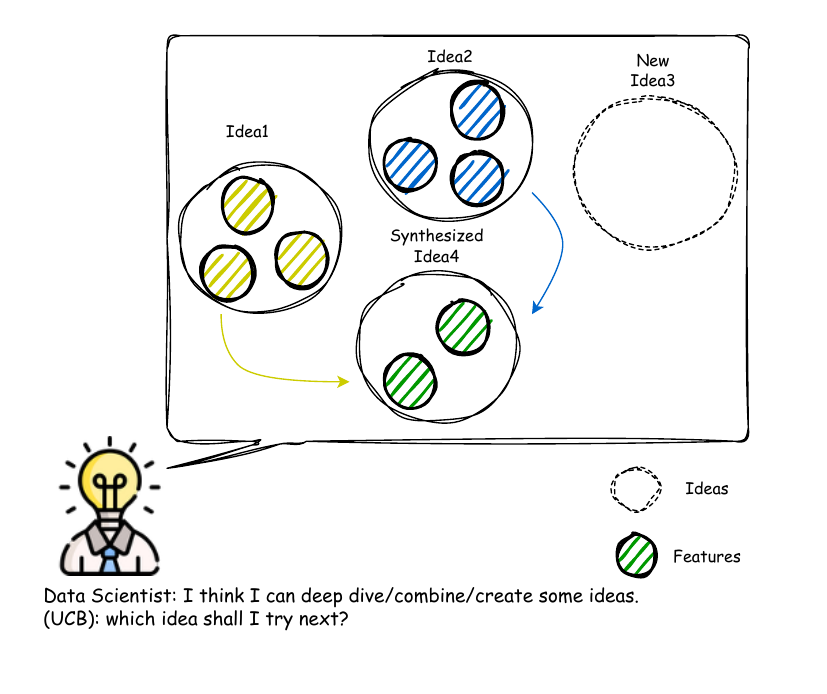}
  \vspace{-8ex}
  \caption{Human data scientists explore existing ideas to derive features, synthesize new concepts, and decide which idea to pursue next through logical reasoning and creative insight.}
  \label{fig:insights}
  \vspace{-4ex}
\end{figure}

\begin{itemize}
    \item \textbf{Multi-agent collaborative architecture.} FELA employs multiple LLM-based agents with specialized roles that collaborate to manage the complexity of automated feature engineering. Leveraging the reasoning capabilities and domain priors of large language models, \textit{idea agents} autonomously generate novel feature hypotheses grounded in dataset semantics. \textit{Code agents} then translate these ideas into executable feature extraction code, enabling expressiveness beyond fixed operation sets used in traditional approaches. Since LLMs are prone to hallucination~\cite{llm-hallucination, liu2024exploring}, particularly in high-dimensional and complex domains, FELA introduces \textit{critic agents} to evaluate and verify outputs from both idea and code agents. This decoupled design separates creativity from verification, effectively mitigating hallucination and improving the reliability of generated features.

    \item \textbf{Explainable and steerable knowledge structure.} FELA organizes knowledge in a hierarchical form consisting of \textit{ideas} and their corresponding derived \textit{features}, as illustrated in Figure~\ref{fig:insights}. Ideas encapsulate high-level, human-readable insights, while features represents their logical and mathematical realizations. This design draws inspiration from cognitive science studies~\cite{cogntive1,cogntive2}, which suggest that human reasoning proceeds top-down—from abstract concepts to more concrete understandings. Unlike existing methods~\cite{llmfe, dynamicllm}, this structure yields a transparent and steerable evolution trajectory, allowing users to control exploration depth and conceptual scope. Moreover, the modular organization facilitates scalability and continual improvement, as new ideas—whether derived from experts or external knowledge sources—can be seamlessly integrated into the system’s knowledge base, enabling ongoing human–machine co-evolution.

    \item \textbf{Agentic evolution algorithm.} We further propose an adaptive learning algorithm that integrates the strengths of genetic algorithms and non-associative reinforcement learning. Idea and feature evolution follows an island-based population paradigm, while a multi-armed bandit mechanism with Upper Confidence Bound (UCB) exploration dynamically balances exploration and exploitation across idea trajectories. To enhance learning efficiency, we employ an \textit{evaluation agent} to summarize experience from past trials and adaptively update the shared system memory. FELA maintains a dual-memory design: short-term memory captures local context for planning and adaptation, whereas long-term memory supports cross-trajectory knowledge transfer. This design allows the system to leverage both evaluation feedback and accumulated historical experience, preventing inefficient wandering in the feature search space.
\end{itemize}

Through these designs, FELA makes the following key academic contributions:
\begin{itemize}
    \item To the best of our knowledge, FELA is the first agentic system capable of performing automated feature engineering on \textit{industrial-scale event log data}.
    \item We introduce a novel hybrid evolution framework that combines genetic and reinforcement learning principles within a hierarchical knowledge structure, enabling efficient and interpretable idea evolution.
    \item Extensive experiments on real-world industrial datasets demonstrate the effectiveness, scalability, and generalizability of our approach.
\end{itemize}

We have deployed FELA in our internal feature engineering platform, saving weeks of labor on manual feature engineering and improving performances of various predictive models across multiple business scenarios.

\section{BACKGROUND \& RELATED WORK}
\subsection{Feature Engineering on Industrial Data}

Feature engineering—also referred to as feature transformation or feature generation—denotes the process of constructing, transforming, and selecting informative variables from raw data to enhance the performance and generalization of machine learning models~\cite{hollmann2024caafe}. By providing models with the most relevant representations, effective feature engineering underpins predictive accuracy, interpretability, and decision-making capability. 

Broadly, existing approaches to feature generation fall into two paradigms. The first, \emph{latent representation learning}, leverages deep architectures such as deep factorization machines and neural representation models to capture complex, nonlinear dependencies in data~\cite{guo2017deepfm, bengio2013representation}. While powerful, these approaches often lack interpretability and traceability~\cite{scholkopf2021causal}. The second paradigm, \emph{explicit feature transformation}, constructs new features through rule-based arithmetic, aggregation, or logical operations~\cite{nargesian2017learning, zhong2016overview}. These methods depend heavily on domain expertise and manual design, which limits their scalability and adaptability across tasks.

In industrial environments, raw data streams exhibit complexities far beyond those of conventional structured or tabular datasets, posing substantially greater challenges for feature engineering. The characteristics of industrial data are often summarized by the ``5Vs'' framework—\emph{Volume, Velocity, Variety, Veracity}, and \emph{Value}~\cite{theodorakopoulos2024state}. Industrial systems typically involve intricate structural dependencies and operational uncertainties, further complicating the extraction of robust and meaningful features. Consequently, there is a growing demand for automated, domain-adaptive feature engineering techniques that can handle heterogeneous, large-scale, and often unstructured data~\cite{hu2019automatic}.

Manual feature engineering, by contrast, is increasingly impractical in such settings. It relies heavily on the tacit domain knowledge of data scientists, who face steep learning curves when interpreting complex raw data. Frequent project transitions further exacerbate the problem, as practitioners must repeatedly invest substantial effort to understand new datasets and devise feature hypotheses from scratch—significantly impairing productivity and slowing innovation.

To mitigate these limitations, recent work has focused on \emph{automated feature engineering (AutoFE)}. Traditional AutoFE methods include tree-based exploration~\cite{khurana2016cognito}, iterative subsampling~\cite{horn2019autofeat}, and transformation enumeration~\cite{kanter2015deep}. Learning-based approaches further exploit machine learning and reinforcement learning to guide the feature construction process~\cite{nargesian2017learning, khurana2018rl, zhang2019deep}. For example, OpenFE~\cite{zhang2023openfe} integrates a feature-boosting algorithm with a two-stage pruning mechanism to achieve expert-level performance. Despite these advances, most existing AutoFE systems struggle to integrate domain expertise and constrained by the predefined operation sets. Recently, large language models (LLMs) have shown promise in this direction, offering contextual reasoning and domain-aware knowledge transfer for feature discovery~\cite{hollmann2024caafe}.

\subsection{LLMs on data centric tasks}

Recent advances in large language models (LLMs) have demonstrated their ability to leverage pre-trained knowledge for novel tasks through techniques such as prompt engineering and in-context learning, often without additional task-specific training \cite{brown2020few, wei2022chain}. Despite these capabilities, LLMs can produce factually incorrect or inconsistent outputs, motivating research into mechanisms that refine their outputs using feedback \cite{madaan2024selfrefine, haluptzok2022self}. 

To further harness LLMs for optimization, recent approaches integrate them with evolutionary and search-based frameworks. LLMs have been used to perform adaptive mutation and crossover operations within evolutionary algorithms \cite{meyerson2024lmcrossover, lehman2023evolution}, and coupled with evaluators to guide the search process \cite{liu2024llmbo, wu2024survey, lange2024llms}. Such strategies have shown success in areas including prompt optimization \cite{guo2023promptopt, yang2024llmopt}, neural architecture search \cite{chen2024evoprompting, zheng2023nas}, symbolic regression \cite{shojaee2024llmsr}, and discovery of mathematical heuristics \cite{romera2024mathdisc}.

Very recently, several studies have begun exploring the use of large language models (LLMs) for tabular data analysis. CAAFE~\cite{hollmann2024caafe} represents one of the earliest attempts to integrate LLMs into the feature generation loop by providing task-specific contextual descriptions and iteratively adding features in a greedy, single-path manner. In contrast, OCTree~\cite{nam2024optimized} introduces a rule-based mechanism inspired by decision trees, where LLMs are employed to generate and refine feature-construction rules, thus making the search process more interpretable and controllable. Focusing on controllability, other approaches~\cite{dynamicllm,gong2025evolutionary} predefine an operation set, using the LLM to select and combine operations at each iteration to produce new features. However, these methods impose overly rigid constraints on the search space, severely limiting the expressive capacity of LLMs and rendering them unsuitable for the industrial-scale complexity considered in this work.

Focusing on \textit{free coding} capability, FeatLLM~\cite{featllm} leverages few-shot prompting to guide LLMs in generating feature-engineering code directly, while LLM-FE~\cite{llmfe}, inspired by FunSearch~\cite{romera2024mathdisc}, employs an island-based evolutionary framework to manage populations and mutations of generated code. Although these approaches offer greater flexibility and more suitable for heterogeneous data, they often sacrifice steerability and fine-grained control over the evolution trajectory. 

In contrast, our method fully exploits the LLM’s coding and reasoning capabilities to perform arbitrarily sophisticated feature transformations while maintaining controllable evolution through a structured \emph{idea–feature} knowledge architecture. This design enables the generation of comprehensive, robust, and interpretable features, making the approach scalable and applicable to the complexities of real-world industrial environments.


\section{Problem Formulation}
\subsection{Dataset}
Let $\mathcal{D}\in \mathbb{R}^{N\times M}$ denotes a tabular dataset, where $N$ is the number of rows and $M$ the number of columns. Each row $x_i$ is associated with a user $u_i$. In non event log scenarios, $N$ simply equals to the number of users $|\mathcal{U}| = L = N$. And rows $\mathbf{x}$ are often assumed to be independent, identically distributed (i.i.d). An event log dataset records a sequence of user interactions with a system, where each interaction corresponds to a discrete event triggered by user behavior (e.g., click, like, play). Formally, let the event log dataset be denoted as
\[
\mathcal{D}_{\text{log}} = \{ x_i \}_{i=1}^N,
\]
where each event instance $x_i$ is represented as a tuple:
\[
x_i = (u_i, a_i, t_i, \mathbf{c}_i).
\]
Here, $u_i \in \mathcal{U}$ denotes the user identifier, $a_i$ denotes the action type (e.g., click, like, buy), $t_i$ is the timestamp at which the event occurs, and $\mathbf{c}_i \in \mathbb{R}^d$ represents the contextual or system state features associated with the event (such as device information, item metadata, or session context).

Since a user may perform multiple actions over time, the same user $u_i$ can appear in multiple events, leading to $|\mathcal{U}| < N$. The events are typically ordered chronologically according to their timestamps $t_1 \le t_2 \le \cdots \le t_N$. Thus, the event log dataset can be viewed as a temporally ordered multiset of user–action pairs with associated contexts, which obviously break the i.i.d assumption mentioned before. Nonetheless, our solution is able to handle both event and non-event log dataset.

\subsection{Task}
For supervised learning tasks, each user $u_i$ is associated with a corresponding label $y_i$, where $y_i \in \{0, 1, ..., K\}$ for classification tasks with $K$ classes, and $y_i \in \mathbb{R}$ for regression tasks. Given a labeled tabular dataset $\mathcal{D} = (u_i(\mathbf{x}), y_i)_{i=1}^L$ and prediction model $f$ to map from the input feature space $\mathcal{X}$ to its corresponding label space $\mathcal{Y}$, our objective is to find an optimal feature transformation $\theta$, which enhances the performance $\mathcal{R}$ of a predictive model when trained on the transformed input space. Formally, the feature engineering task can be defined as:

\begin{equation}
\label{equ: target}
\max_{\theta} \, \mathcal{R}\big(f^*(\theta(\mathcal{X'})), \mathcal{Y'}\big), \{\mathcal{X'}, \mathcal{Y'}\} \sim \mathcal{D}_{test} 
\end{equation}
subject to:
\begin{equation}
f^* = \arg\min_{f} \, \mathcal{L}_f\big(f(\theta(\mathcal{X})), \mathcal{Y}\big), \{\mathcal{X}, \mathcal{Y}\} \sim \mathcal{D}_{train} 
\end{equation}
where $\mathcal{D}_{train}$ and $\mathcal{D}_{test}$ are the training set and test set, respectively. 


\subsection{LLM Agents}
We now introduce the large language model (LLM) agents that collaboratively construct the optimal feature transformation $\theta$. Each agent $\mathcal{A}_p$ is formulated as a sequence-to-sequence function that operates on textual or structured representations of the current system state. The overall feature transformation $\theta$ is achieved through the sequential collaboration of $P$ agents:
\[
\theta = \mathcal{A}_1 \circ ...\circ  \mathcal{A}_{P-1} \circ \mathcal{A}_{P}
\] 
through manipulating the data definition and prior knowledge.

\begin{figure*}[h]
  \includegraphics[width=\textwidth]{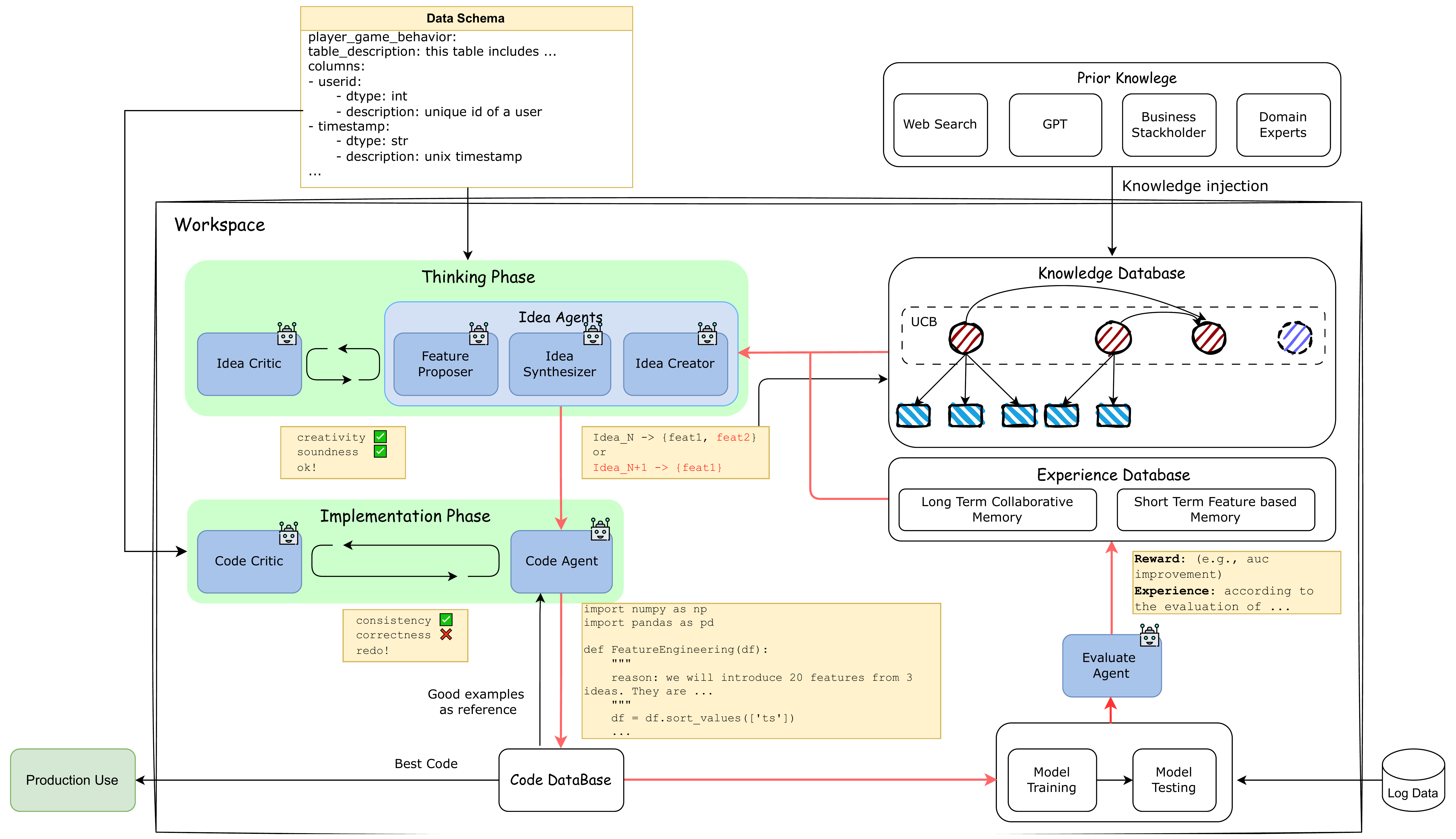}
  \caption{\textbf{System overview of FELA.} The red arrows illustrate the core self-evolution loop of the system. The \emph{idea agents} propose new feature concepts or experimental ideas, which are then translated into executable feature engineering code by the \emph{code agents}. The generated code is executed and evaluated by the \emph{evaluate agent} using real event log data to produce corresponding performance rewards. The resulting experiences are summarized and updated into both the short-term memory and the global long-term memory, guiding subsequent rounds of exploration. The \emph{idea critic} and \emph{code critic} modules enhance robustness and consistency by validating generated ideas and code, respectively. The best-performing features and implementations can be seamlessly exported to production environments.}
  \label{fig:system}
\end{figure*}
\section{Methodology}

\subsection{System Overview}
Figure~\ref{fig:system} illustrates the overall architecture of \textbf{FELA}, which consists of two major components: (1) external dependencies, including the data schema, prior knowledge, and log data\footnote{Without loss of generality, we use log data in the following discussion. However, our system can be easily applied to simpler tabular datasets.}; and (2) the internal agent workspace.

\textbf{External Dependencies.} The \emph{data schema} $\mathcal{H}$ provides metadata describing the dataset, such as the context of data collection, column types, and textual descriptions. This information grounds the reasoning process and mitigates potential hallucinations by offering sufficient contextual grounding for large language model (LLM) agents. The \emph{prior knowledge} component captures textual insights gathered from domain experts, stakeholder interviews, or external web resources. These insights are stored in a knowledge base to guide the system’s initial exploration. Although FELA can operate without explicit prior knowledge, such injections enable stronger control over the search space and alignment with domain-specific objectives. Finally, the \emph{event log data} serves as the operational environment and provides the material basis for generating experimental results and evaluating newly proposed features.

\textbf{Internal Agent Workspace.} Within the agent workspace, FELA employs a team-based agent framework that iteratively generates, critiques, implements, and evaluates new feature ideas. The \emph{idea agents} consider the data schema, current knowledge base, and past experience to propose plausible new features. Each proposal is subsequently reviewed by an \emph{idea critic}, which evaluates its novelty, soundness, and adherence to predefined quality criteria. Approved ideas are then passed to the \emph{code agents}, which translate high-level feature concepts into executable feature engineering code. The generated code is validated by a \emph{code critic} to ensure syntactic correctness and logical consistency before being added to the code repository.

The approved code is executed to transform the raw feature space, and the resulting features are used in downstream model training and evaluation under the target performance metrics (e.g., accuracy, AUC-score). The evaluation results are then processed by an \emph{evaluate agent}, responsible for two key functions: (1) updating the performance attributes of the evaluated idea and corresponding feature implementations (2) synthesizing higher-level experience abstraction as global long-term memory. The updated knowledge base and memory modules subsequently inform the next round of idea generation, enabling FELA to perform self-improving, closed-loop evolution in feature engineering.

In the following sections, we detail the agent interaction protocol, the reward formulation for idea evaluation, and the mechanisms enabling FELA’s long-term evolution and knowledge retention.






    


\begin{algorithm}[t]
\caption{\textbf{Main Evolution Loop of FELA}}
\label{alg:fela}
\KwIn{Data schema $\mathcal{S}$, prior knowledge $\mathcal{K}_0$, event log data $\mathcal{D}$, target metrics $\mathcal{R}$}
\KwOut{Optimized feature set $\mathcal{F}^*$, best code $\theta^*$}

Initialize short-term memory $\mathcal{M}_s$, long-term memory $\mathcal{M}_l$, and knowledge base $\mathcal{K} \leftarrow \mathcal{K}_0$\;
Initialize empty feature set $\mathcal{F} \leftarrow \emptyset$ and code set $\mathcal{C} \leftarrow \emptyset$\;

\While{not converged}{
    \tcp{--- Idea generation stage ---}
    \While{\text{IdeaCritic(idea)} == \texttt{reject}}{
        $\text{idea} \leftarrow \text{IdeaAgent}(\mathcal{S}, \mathcal{K}, \mathcal{M}_s, \mathcal{M}_l)$\;
    }

    \tcp{--- Implementation stage ---}
    \While{\text{CodeCritic(code)} == \texttt{reject}}{
        $\text{code}~\theta  \leftarrow \text{CodeAgent}(\text{idea}, \mathcal{S})$\;
    }

    \tcp{--- Evaluation stage ---}
    $\text{features} \leftarrow \text{Execute(code}~\theta,  \mathcal{D}\text{)}$\;
    $\text{reward} \leftarrow \text{Evaluate(features, } \mathcal{R}\text{)}$\;

    \tcp{--- Learning and evolution stage ---}
    $\mathcal{K} \leftarrow \text{EvaluateAgent}(\text{idea}, \text{reward})$\;
    $\mathcal{M}_l \leftarrow \text{EvaluateAgent}(\mathcal{M}_l, \text{idea}, \text{reward})$\;

    \tcp{--- Selection and refinement ---}
    \text{UpdateExplorationPolicy(UCB, reward)}\;
    $\mathcal{F} \leftarrow \mathcal{F} \cup \text{features}$\;
    $\mathcal{C} \leftarrow \mathcal{C} \cup \theta $\;
}

$(\mathcal{F}^*, \theta^*) \leftarrow \text{SelectBestSolution}(\mathcal{F}, \mathcal{C}, \mathcal{R})$\;
\Return $(\mathcal{F}^*, \theta^*)$\;
\end{algorithm}

\subsection{Knowledge Base}
\begin{figure}
  \includegraphics[width=0.52\textwidth]{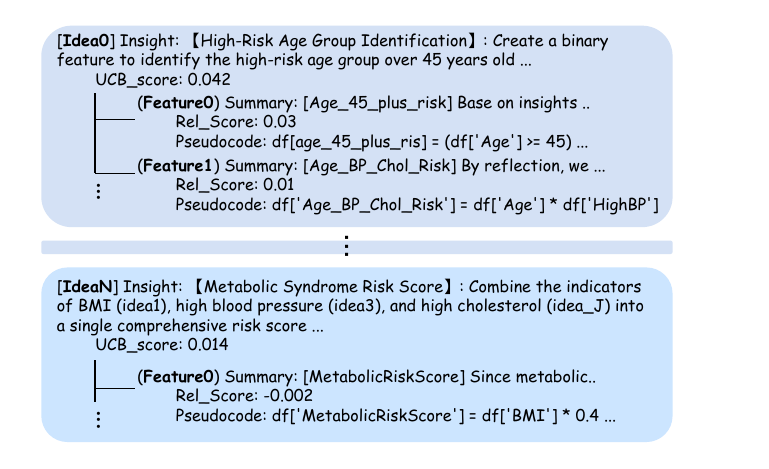}
  \caption{\textbf{An illustration of the knowledge base.} It contains rich information including idea insights, ucb scores, associated feature and corresponding pseudocodes, etc. }
  \label{fig:knowledge_base}
  \vspace{-5pt}
\end{figure}
The knowledge base serves as the central repository of textual insights and their corresponding feature implementations. Formally, it is defined as $\mathcal{K} = \{I_i\}_{i=1}^{N_I}$, where each idea $I_i$ represents an abstract understanding of the underlying data scenario. For example, in an e-commerce advertising context, an insight might be expressed as “Users tend to purchase complementary items within the same session.” Although such an idea does not directly specify how to compute a feature, it encapsulates a meaningful intuition—analogous to how human data scientists begin their exploratory reasoning in practice.

Each idea $I_i$ is associated with one or more features ${d_{ij}}$, which provide concrete realizations of how the insight can be operationalized logically and mathematically. In the above example, “complementary items” can be interpreted in multiple ways—such as categorical complementarity (e.g., “iPhone” and “iPhone case”) or temporal complementarity (e.g., “diapers” and “beer” purchased sequentially). These distinct interpretations correspond to different features of the same high-level insight. Together, the ideas and their features form a two-layer hierarchical structure, as illustrated in Figure~\ref{fig:knowledge_base}, representing the conceptual-to-operational flow of feature construction. Formally, the two-layer hierarchical knowledge base $\mathcal{K}$ can be expressed as 
\[
\mathcal{K} = \{ (I_i, \{ d_{i,j} \}_{j=1}^{M_i}) \}_{i=1}^{N_I},
\]
where $M_{i}$ denotes the number of features in $i$th idea. Drawing an analogy to island-based genetic algorithms, each idea can be viewed as an “island,” while its associated features constitute the “population” of that island. Features within the same island may undergo mutation to explore local variations, while different islands can interact and exchange information to generate novel ideas—a process that will be elaborated in the following section.

\subsection{Idea Agents}
Operating in the hypothetical space, the role of \textit{idea agents} is to expand the knowledge base and drive its evolution. At each iteration step $t$, an idea agent takes as input the current knowledge database $\mathcal{K}_t$ and selects one of the following actions: (1) generate a new feature from an existing insight, (2) synthesize a new insight from existing insights, or (3) create an entirely new idea. Each action is carried out by a specialized sub-agent.

\textbf{Feature Proposer.} When option (1) is chosen, we employ an Upper Confidence Bound (UCB) algorithm to sample an idea for exploitation and generate a new feature. The details of UCB algorithm are provided in Sec.~\ref{subsec: learning algorithm}. For a given idea, we retrieve its associated past experiences to construct the short-term memory for that idea. Based on this memory $\mathcal{M}_s$ and the adaptively improved global long-term memory $\mathcal{M}_l$, the LLM-based agent proposes a new feature implementation that is (i) consistent with the abstract insight, (ii) technically sound, and (iii) creative—i.e., distinct from existing implementations under the same idea. Formally,
\begin{equation}
    d_{i,j+1} = \mathcal{A}_{\text{feature}}(\mathcal{M}_{s}, \mathcal{M}_{l}, I_{i}, \mathcal{H})
\end{equation}

The feature implementation $d$ is represented as a tuple $(\text{reason}, \text{summary}, \text{pseudocode})$. The \emph{reason} field allows the idea agent to explicitly articulate the rationale behind proposing a specific implementation. The \emph{summary} provides a concise description of the feature's underlying intuition. The \emph{pseudocode} encodes the programmable instructions for the downstream code agent, including all necessary dependencies to generate the targeted feature. Collectively, this tuple establishes a seamless pathway from high-level abstract insight to detailed, implementable instructions. We find that such a structured representation is critical for ensuring robustness and consistency in the subsequent code generation process.

\begin{figure}
  \includegraphics[width=0.48\textwidth]{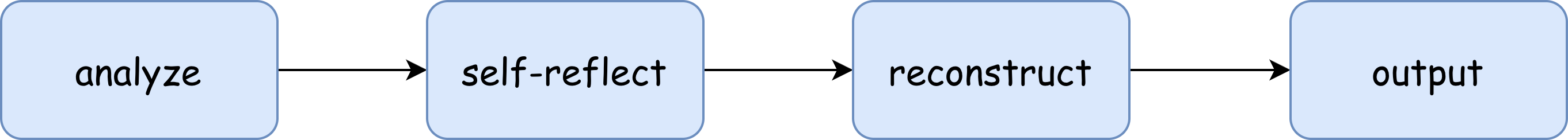}
  \caption{\textbf{Thinking paradigm of our agents.} All agents follow this reasoning framework unless otherwise stated.}
  \label{fig:think_paradigm}
  \vspace{-10pt}
\end{figure}

To ensure that the generated feature implementations satisfy these criteria, we adopt the reasoning spirit of ReAct~\cite{yao2022react} and Reflexion~\cite{shinn2023reflexion}, and employ a three-step reasoning framework (see Figure~\ref{fig:think_paradigm}) for our agents. Specifically, the first step, \textit{analyze}, gathers all relevant information, examines the current state, and produces a textual analysis trace. This step encourages the LLM to form a deep understanding of the provided context. The second step, \textit{self-reflect}, serves as a self-critique phase, prompting the model to check for factual errors or inconsistencies against the task requirements. The final step, \textit{reconstruct}, allows the LLM to revise its prior reasoning based on the self-reflection results, thereby fostering deeper understanding and improved reasoning before producing the final \textit{output}.

\textbf{Idea Synthesizer.} When option (2) is selected, the LLM is instructed to synthesize new ideas by combining existing ones, guided by the current ideas and their associated evaluation scores. This process mimics how humans interconnect related concepts to generate novel insights. While prior work typically selects which ideas (e.g., two) may be combined, we observe that LLMs are inherently capable reasoners that can exploit contextual cues to form meaningful combinations—thanks to the strong priors embedded in their parameters. Therefore, we provide the model with all existing ideas in the knowledge base $\mathcal{K}$ (including newly generated ones) and instruct it to propose a new idea satisfying three criteria: novelty, soundness, and consistency. Formally,
\begin{equation}
    I_{i+1} = \mathcal{A}_{\text{synthesize}}(\mathcal{M}_l, \mathcal{K}, \mathcal{H})
\end{equation}
During the analysis phase, the LLM is explicitly required to ground its innovation in existing ideas, which, according to our observations, effectively reduces hallucination.

\textbf{Idea Creator.} The \textit{idea creator} is responsible for generating entirely new ideas that expand the knowledge base beyond previously injected prior knowledge. Taking the entire knowledge base $\mathcal{K}$ as input, and leveraging the LLM’s strong reasoning and abstraction capabilities, the idea creator can propose novel, logically sound, and schema-consistent ideas that have not been represented before. This capability stems from the extensive corpus used during LLM pre-training, which implicitly encodes a vast reservoir of latent knowledge across diverse domains. By systematically eliciting this embedded intelligence and encouraging creative reasoning, the idea creator continuously enriches the knowledge base with valuable new insights.\footnote{In fact, we can even enroll the updated knowledge from internet such as WebAgent~\cite{gur2024real}. However, to maintain controlled experiments, our experiment setup restricts agents to offline operation.} Formally,
\begin{equation}
    I_{i+1} = \mathcal{A}_{\text{create}}(\mathcal{M}_l, \mathcal{K}, \mathcal{H}).
\end{equation}

\subsection{Code Agent}

The role of the code agent is to faithfully translate the feature implementation pseudocode and associated analysis into executable feature engineering code. Given a feature implementation produced by the idea agent, along with high-quality examples from previous attempts and the data schema as concrete guidance, the code agent generates highly executable and arbitrarily sophisticated code $\theta_t$ to transform raw features:

\begin{equation}
    \theta_{t} = \mathcal{A}_{\text{code}}(\{\{d_{i,j}\}_{j=0}^{M_{i}}\}_{i=0}^{k}, \mathcal{H}, \theta_{j<t}),
    \label{code_agent_equation}
\end{equation}
where $\{\{d_{i,j}\}_{j=0}^{M_{i}}\}_{i=0}^{k}$ denotes selected feature implementations from $k$ ideas, and $t$ denotes the iteration step in the FELA system.

Compared to existing LLM-based approaches, our solution emphasizes two key improvements. First, instead of relying on a predefined feature operation set as in \cite{dynamicllm, tang2025alphaagent}, we allow the LLM to code freely. While operation sets provide high controllability, they are extremely limited and inefficient for two reasons: (i) many feature computations cannot be represented by simple binary operations (e.g., k-means clustering), and complex features often require multiple rounds of refinement; (ii) exhaustively enumerating all possible arithmetic operations is practically infeasible, which imposes an upper bound on the expressiveness of feature engineering.  

Second, unlike prior free-coding approaches~\cite{llmfe}, which often produce inconsistent results and degrade in quality as context length increases over iterations, our method grounds code generation on the underlying ideas and feature implementations. These implementations provide precisely the right context to guide the LLM coder, ensuring robust and steerable growth of feature engineering code. This design enables FELA to generate dozens of new features spanning thousands of lines of code, while maintaining consistency, correctness, and interpretability.

\subsection{Critic Agents}

Large Language Models (LLMs) often suffer from \emph{attention distillation} issues~\cite{ge2025introducing, hong2025context} when dealing with long contexts, making them prone to hallucinations and reasoning failures. This challenge is particularly severe in our setting, where both ideas and their implementations evolve into large and complex structures over time. In such cases, idea agents and code agents may deviate from intended instructions or produce low-quality results.  

To address this problem, we introduce two auxiliary agents—\textit{idea critics} and \textit{code critics}—which follow a unified design philosophy. Given the output from an idea agent (or code agent), a critic agent is responsible for rigorously evaluating whether the newly generated idea (or code) satisfies the requirements of \textbf{consistency} and \textbf{correctness}. Consistency refers to the alignment with the data schema and the existing knowledge base, while correctness ensures that the proposed idea or code is reasonable and technically sound.  

If any requirement is violated, the critic agent provides detailed feedback to the corresponding generator (idea or code agent), allowing it to revise and improve its output before proceeding. The incorporation of these critics is crucial for maintaining the robustness and reliability of a multi-agent system operating over extended periods—an essential property for real-world industrial applications.  

Formally, the critic’s operation can be represented as:
\begin{equation}
    \mathcal{Z}_t = \mathcal{A}_{\text{critic}}(\{\{d_{i,j}\}_{j=0}^{M_{i}}\}_{i=0}^{k}, \mathcal{H}, C)
\end{equation}
where $\mathcal{Z}_t$ denotes the critic feedback and $C$ represents the set of evaluation criteria. Consequently, the update equation for the code agent (Eq.~\ref{code_agent_equation}) becomes:
\begin{equation}
    \theta_{t, z} = \mathcal{A}_{\text{code}}(\{\{d_{i,j}\}_{j=0}^{M_{i}}\}_{i=0}^{k}, \mathcal{H}, \mathcal{Z}_{z-1})
    \label{code_agent_equation}
\end{equation}
In practice, we define a maximum number of critic iterations $Z$. When $z > Z$, the system forfeits the current refinement loop and initiates a new round of exploration. Similar critic feedbacks are applied to idea agents.

\subsection{Evaluate Agent}

After each round of model training and testing, we obtain feedback signals (e.g., AUC scores) from the downstream prediction task. However, such feedback is extremely sparse relative to the vast—if not infinite—search space in our system, where ideas, implementations, and codes all require exploration and reward-driven updates. Traditional AutoML systems often operate under either a limited search space or a setting where Monte Carlo sampling of candidate actions is computationally cheap. In contrast, in our LLM-based framework, each action (e.g., generating a new idea or implementation) is both computationally and semantically expensive.  

To overcome this challenge, we leverage the intrinsic reasoning capability of LLMs to perform an additional step of reflective analysis after receiving reward feedback. The evaluation agent interprets the reward signal and produces an informative summary of the current exploration trajectory and its corresponding outcomes. This process effectively increases the \emph{bandwidth of the reward channel}, enabling more targeted and precise search in subsequent iterations.  

All such analyses are persistently maintained through our long–short term memory system, which captures both local (short-term) contextual feedback and global (long-term) strategic insights. The details of this memory mechanism are discussed in the following section. Formally, we have 
\begin{equation}
    \mathcal{M}_l^{t+1} = \mathcal{A}_{eval}(\mathcal{I}_t,d_t, \mathcal{R}_t, \mathcal{M}_l^{t})
\end{equation}

\begin{figure}[h]
    \centering
    \includegraphics[width=0.4\linewidth]{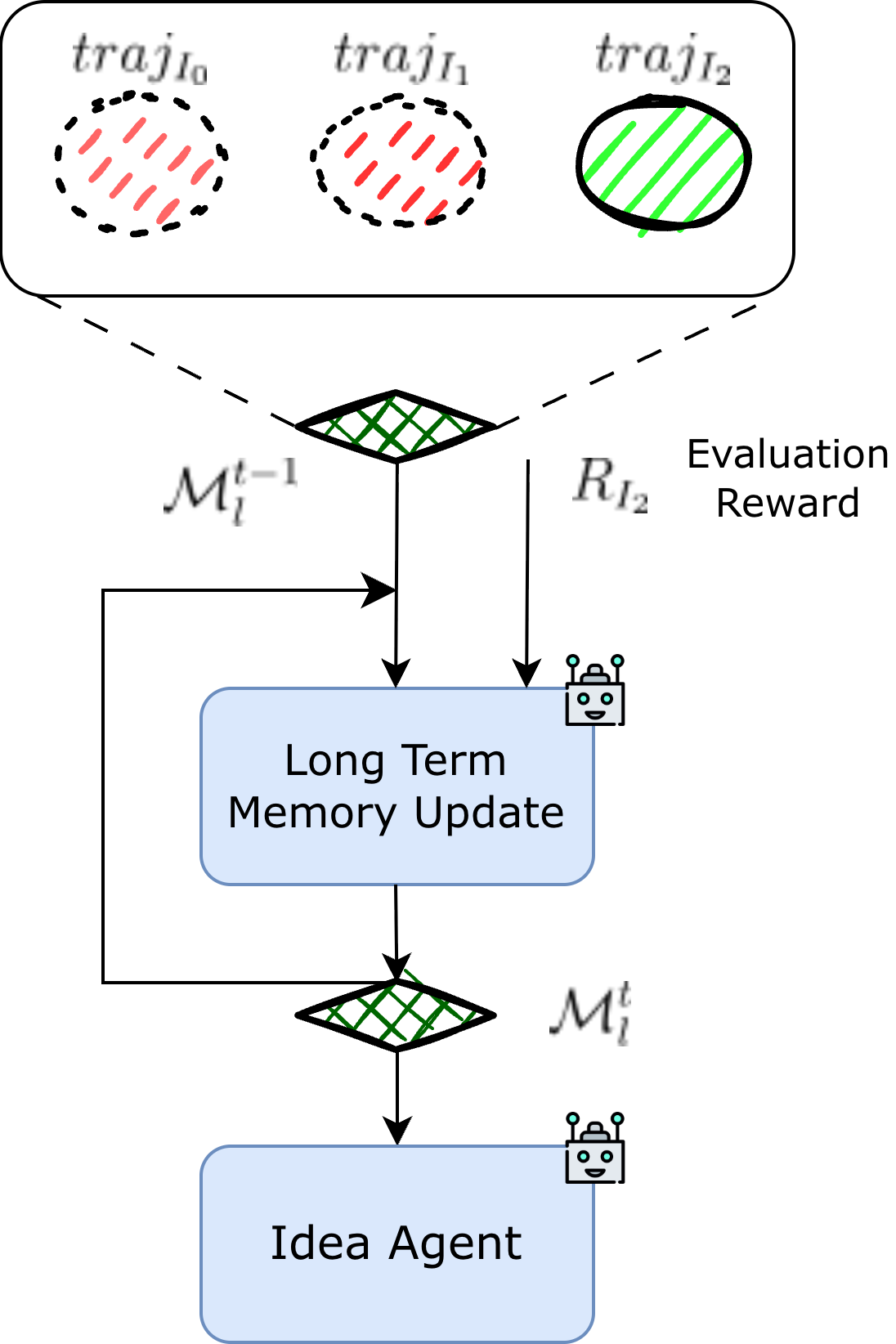}
    \caption{Long Term Memory. Long term memory is updated in an adaptive manner along with the evolution iterations.}
    \label{fig:long_term_mem}
    \vspace{-15pt}
\end{figure}
\subsection{Memory System}
The memory system in FELA is designed to harness past evolutionary experiences to guide future exploration effectively. It comprises two distinct components: a long-term collaborative memory and a short-term feature-based memory.

The long-term memory (see Figure~\ref{fig:long_term_mem}), denoted as $\mathcal{M}_{l}$, captures high-level, natural language expolation insights that are shared across the entire knowledge base. Following each exploration cycle, a dedicated long-term memory agent progressively updates $\mathcal{M}_l$ by incorporating the current state of the evolving idea and its associated rewards. This refreshed long-term memory is then utilized to inform the idea-generation process in the subsequent exploration round.
\begin{figure}[b]
    \centering
    \includegraphics[width=0.8\linewidth]{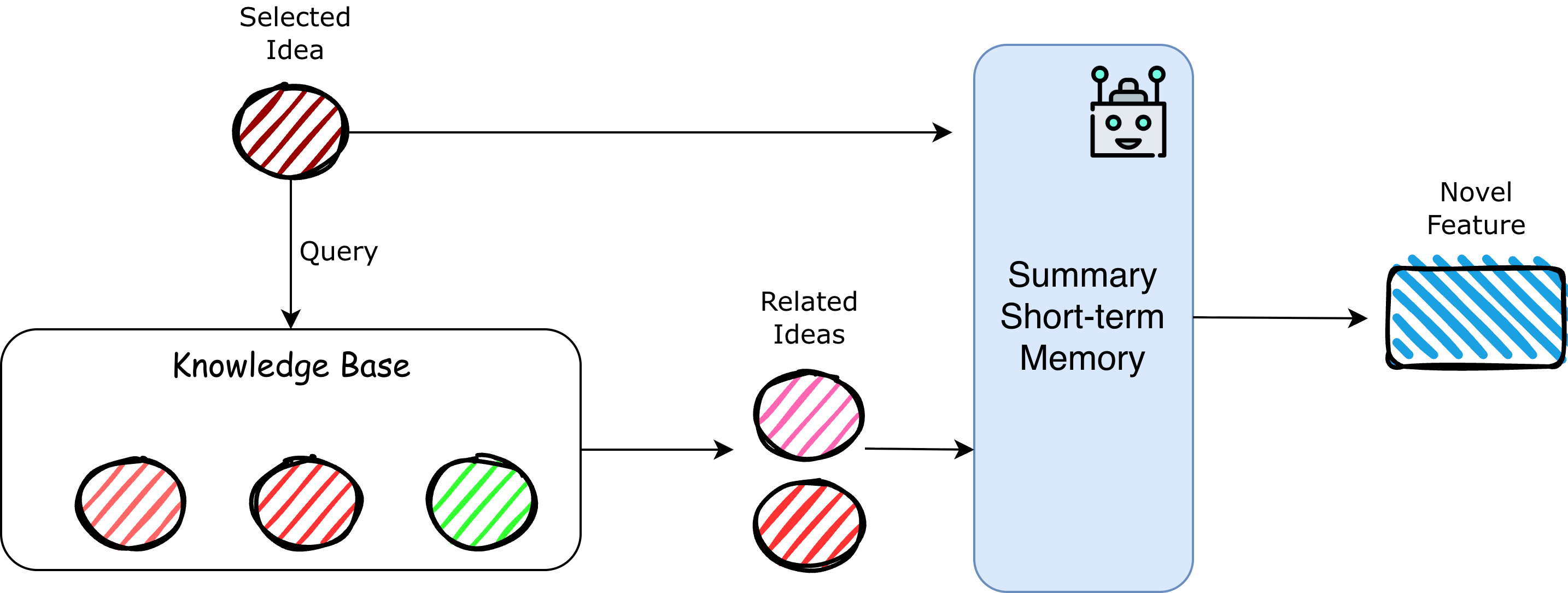}
    \caption{Short Term Memory. The related ideas are retrieved in using RAG.}
    \label{fig:short_term_mem}
\end{figure}

The short-term memory (see Figure~\ref{fig:short_term_mem}), denoted as $\mathcal{M}_s$, provides specific, actionable suggestions for augmenting the current idea with new features, which operates in a retrieval-augmented generation (RAG) manner. To generate $\mathcal{M}_s$, the system first retrieves a set of ideas semantically related to the current one. This is achieved by embedding each idea's insight and feature set into a vector space, where the top-$K$ most similar ideas are retrieved based on cosine similarity. Then, the current idea and its retrieved neighbors are processed by an LLM to produce the short-term memory $\mathcal{M}_s$. This output is fed into the idea agent, which leverages the positive and negative features from related ideas to make more informed and effective modifications, thereby steering the evolutionary process more efficiently.

\subsection{Learning Algorithm}
\label{subsec: learning algorithm}
FELA introduces a novel evolutionary learning algorithm that integrates RL and GA to enhance feature evolution efficiency. To quantify the utility of a newly generated feature $d_{i,j+1}$ within idea $I_i$, we define a relative score $s(d_{i,j+1})$ as
\begin{equation}
\label{equ:relative_score}
s(d_{i,j+1}) = \mathcal{R}\left(f^*(\theta_{i,j+1}(\mathcal{X'})), \mathcal{Y'}\right) - \mathcal{R}\left(f^*(\theta_{i,j}(\mathcal{X'})), \mathcal{Y'}\right),
\end{equation}
where $\theta_{i,j}$ and $\theta_{i,j+1}$ denote the feature transformation functions applied to the feature sets $\{d_{i,j^\prime}\}_{j^\prime=1}^{j}$ and $\{d_{i,j^\prime}\}_{j^\prime=1}^{j+1}$, respectively. This score directly measures the additional contribution of $d_{i,j+1}$ to the predictive performance of $I_i$. A positive $s(d_{i,j+1})$ indicates a beneficial feature, which is retained for subsequent code generation; conversely, a negative score leads to the feature being recorded as a negative sample for the idea. To balance exploration and exploitation during evolution, FELA employs the Upper Confidence Bound (UCB) criterion to select the idea for the next evolution round. The UCB for idea $I_i$ is defined as
\begin{equation}
\label{equ:ucb}
\text{UCB}(I_i) = \frac{\sum_{j} s(d_{i,j})}{Q_i} + c \sqrt{\frac{\log Q}{Q_i}},
\end{equation}
where $Q_i$ is the visit count of $I_i$, $Q$ is the total visits across all ideas, and $c$ is a hyperparameter controlling exploration. The idea with the highest UCB is selected for further evolution, effectively balancing the exploration of new feature combinations and the exploitation of high-performing ideas.

\begin{table}[t]
\caption{Summary of Experimental Datasets}
\label{tab:datasets}
\centering
\begin{tabular}{l l c c l c}
\toprule
\textbf{Dataset} & \textbf{Type} & \textbf{Instances/Records} & \textbf{Features} & \textbf{Classes} \\
\midrule
Dia & Tabular & 253,680 & 21 & 2 \\
Taobao & Log & 420,721 & 31  & 2 \\
Tencent & Log & 630,069 & 23 & 2 \\
\bottomrule
\end{tabular}
\vspace{0.1cm}
\end{table}

\section{Experiment}
\subsection{Experiment Overview}
In this section, comprehensive experiments are conducted to validate the effectiveness and robustness of the proposed FELA framework. Our evaluation is designed to answer the following key questions:
\begin{itemize}
\item \textbf{RQ1 (Performance):} Can FELA outperform raw features and state-of-the-art LLM-based feature engineering methods on tabular and industrial log data?
\item \textbf{RQ2 (Interpretability):} Does FELA provide a steerable and semantically meaningful feature generation process?
\item \textbf{RQ3 (Compatibility):} Is FELA compatible with various machine learning models when integrated within the evolution loop?
\item \textbf{RQ4 (Mechanism):} How does FELA's core design, including the critic agent, long-short-term memory, and the UCB-based learning, contribute to its overall performance?
\item {\textbf{RQ5 (Sensitivity): }How sensitive is FELA over the LLM models and hyper-parameter setting?}
\end{itemize}
To this end, we perform extensive comparisons on real-world datasets, an applicability test across multiple classifiers, a detailed ablation study, and in-depth case analyses on feature correlation and evolution trajectory. The results collectively demonstrate that FELA establishes a new state-of-the-art agentic feature engineering.
\subsection{Experiment Setup}
\label{subsec: setup}
\subsubsection{Datasets and Feature Engineering Task} In the experiment, three real-world datasets are adopted for evaluation, including Diabetes Health Indicator Dataset (Dia) \cite{teboul2022diabetes}, Tabao Conversion Prediction Data (Taobao) \cite{taobao2022ad}, and the User Churn Data in Tencent Game Platform (Tencent). The basic information of the three datasets is summarized in Table~\ref{tab:datasets}, where Dia is standard tabular data while the others are event log data with real industrial complexity.
\begin{itemize}
    \item \textbf{Dia Dataset}: This tabular dataset contains 253,680 instances, where each instance has 21 features and is labeled with a binary class. The feature engineering functions are independently applied to each instance in the dataset, where the output features are used for downstream classification. Here, we split 55\% of the samples for ML model training, and the remaining 45\% of the samples are used for validation.
    \item \textbf{Taobao Dataset}: This industrial log dataset contains 176,773 users, 9599 items, and 420,721 ad records over 7 days, where each record is labeled with a binary ad impression. The data is organized across five relational tables containing basic identifiers, item profiles, user demographics, contextual signals, and shop reputation metrics, which constitute 31 features for each record. The feature engineering functions jointly process the entire log data and output the features for each user-ad pair. Following standard temporal evaluation protocols, the dataset is chronologically split with the initial 6 days for training and the final day for testing, effectively simulating real-world conversion prediction scenarios where models forecast future conversion behavior based on learned historical patterns. \footnote{{The data preprocessing code and initial ideas are released at https://github.com/DKhaoyu/FELA-Feature-Engineering.}}
    \item \textbf{Tencent Dataset}: This dataset is collected from a Massively Multiplayer Online (MMO) game in Tencent for a few days, where the players are sampled from the same user profile to satisfy the i.i.d assumption. Each player is associated with a churn label, indicating whether the user will churn in the next day. In our scenario, each player can have multiple game plays in a day, which results in multiple event log instances. Furthermore, each player in the same game can have multiple teammates, and if the player has more than one teammate, we will separate the records into different rows. Therefore, the instances in such event log data do not follow the i.i.d assumption as mentioned in the introduction session. The feature engineering functions jointly process the entire log data and output the features for each player. The downstream churn prediction is thus working on the player level. For evaluation, a temporal split is adopted: the former day of data are used for training, and the logs from the latter day are held out for testing.

\end{itemize}
\subsubsection{Baselines}
We evaluate the performance of our proposed FELA system against the following baselines: {(1) raw features, (2)  OpenFE \cite{zhang2023openfe}, (3) skope-rules \cite{skope}, (4) CAAFE \cite{hollmann2023large}, (5) FeatLLM \cite{han2024large}, (6) AIDE \cite{jiang2025aide}, (7) LLM-FE \cite{llmfe}. Here,  skope-rules is a traditional rule induction method, the OpenFE is the latest baseline in conventional AutoFE, while CAFFE, FeatLLM, AIDE, and LLM-FE are state-of-the-art LLM-based free coding methods.} To ensure a fair comparison, all LLM-based methods, including the baselines and our proposed FELA, utilize the DeepSeek-V3 model via the DeepSeek API by default \cite{liu2024deepseek}. 
\subsubsection{Metrics}
The feature engineering for FELA and the LLM-FE baseline is optimized using the AUC-score, which we denote as the target metric \(\mathcal{R}\) in Equation~\eqref{equ: target}. The generated features of all methods are then evaluated using a Random Forest classifier by default. { The performance is reported across the main AUC metric and four auxiliary metrics, including Accuracy, Precision, Recall, and F1-score. }

\begin{table}[!t]
\caption{{Performance comparison on different datasets.}} 
\label{tab:datasets}
\centering
\begin{tabular}{lcccccc}
\toprule
\textbf{Dataset} & \textbf{Method} & \textbf{Acc} & \textbf{Prec} & \textbf{Recall} & \textbf{F1} & \textbf{AUC} \\
\midrule
\multicolumn{1}{c}{\multirow{8}{*}{\centering Dia}} & Raw Features & 0.859 & 0.489 & 0.166 &0.247 & 0.797 \\
& Skope-rules & 0.858 & 0.482 & 0.171 & 0.253 & 0.797 \\
& OpenFE & 0.858 & 0.484 & 0.185 & 0.268 & 0.795 \\
& CAAFE & 0.860 & 0.499 & 0.170 & 0.254 & 0.799 \\
& AIDE & 0.859 & 0.489 & 0.166 & 0.247 & 0.797 \\
& FeatLLM & 0.862 & 0.489 & 0.093 & 0.159 & 0.797 \\
& LLM-FE & 0.861 & 0.514 & 0.155 & 0.239 & \underline{0.802} \\
& FELA (Ours) & 0.863 & 0.552 & 0.149 & 0.235 & \textbf{0.812} \\
\midrule
\multicolumn{1}{c}{\multirow{8}{*}{\centering Taobao}} & Raw Features & 0.630 & 0.0269 & 0.580 & 0.0514 & 0.630 \\
& Skope-rules & 0.660 & 0.0279 & 0.551  & 0.0531 & 0.639 \\
& OpenFE & 0.592  & 0.0264 & 0.630 & 0.0506 & 0.636  \\
& CAAFE & 0.554 & 0.0243 & 0.638 & 0.0471 & 0.622 \\
& AIDE & 0.630 & 0.0264 & 0.569 & 0.0504 & 0.628 \\
& FeatLLM & 0.645 & 0.0260 & 0.536 & 0.0496 & 0.631 \\
& LLM-FE & 0.678 & 0.0286 & 0.535 & 0.0543 & \underline{0.641} \\
& FELA (Ours) & 0.678 & 0.0294 & 0.551 & 0.0558 & \textbf{0.653} \\
\midrule
\multicolumn{1}{c}{\multirow{8}{*}{\centering Tencent}} & Raw Features & 0.668 & 0.576 & 0.385 & 0.461 & 0.683 \\
& Skope-rules & 0.674 & 0.595 & 0.367 & 0.454 & 0.685 \\
& OpenFE & 0.672 & 0.578 & 0.418 & 0.485 & 0.686 \\
& CAAFE & 0.671 & 0.581 & 0.394 & 0.470 & 0.686 \\
& AIDE & 0.668 & 0.576 & 0.385 & 0.461 & 0.683 \\
& FeatLLM & 0.649 & 0.538 & 0.362 & 0.433 & 0.657 \\
& LLM-FE & 0.672 & 0.583 & 0.395 & 0.471 & \underline{0.686} \\
& FELA (Ours) & 0.680 & 0.604 & 0.387 & 0.472 & \textbf{0.701} \\
\bottomrule
\end{tabular}
\vspace{-10pt}
\end{table}

\subsection{Feature Engineering Performance (RQ1)}
Table~\ref{tab:datasets} presents the feature engineering performance of FELA and baseline methods on different datasets. In the Dia dataset, the proposed FELA achieves superior performance in accuracy, precision, and AUC, demonstrating its effectiveness in handling structured data. Notably, FELA improves the AUC from 0.802 to 0.812 compared to the runner-up method LLM-FE, indicating its capability to generate high-quality discriminative features for tabular data analysis. On the Taobao conversion prediction dataset, FELA maintains its competitive advantage. Our method obtains a notable AUC improvement from 0.641 to 0.653 over LLM-FE. This consistent superiority across multiple metrics confirms FELA's robustness in processing complex industrial log data with relational characteristics. The evaluation on the Tencent user churn dataset further validates FELA's superiority on complex industrial log data. FELA outperforms all baselines, achieving a significant AUC boost from 0.686 to 0.701 compared to LLM-FE {and OpenFE}. These results across three distinct domains demonstrate that FELA consistently generates semantically meaningful features that enhance predictive performance in diverse real-world industrial scenarios. 

\subsection{Steerable Feature Generation (RQ2)}
\label{subsec: case study}
\begin{figure*}
    \centering
    \includegraphics[width=0.95\linewidth]{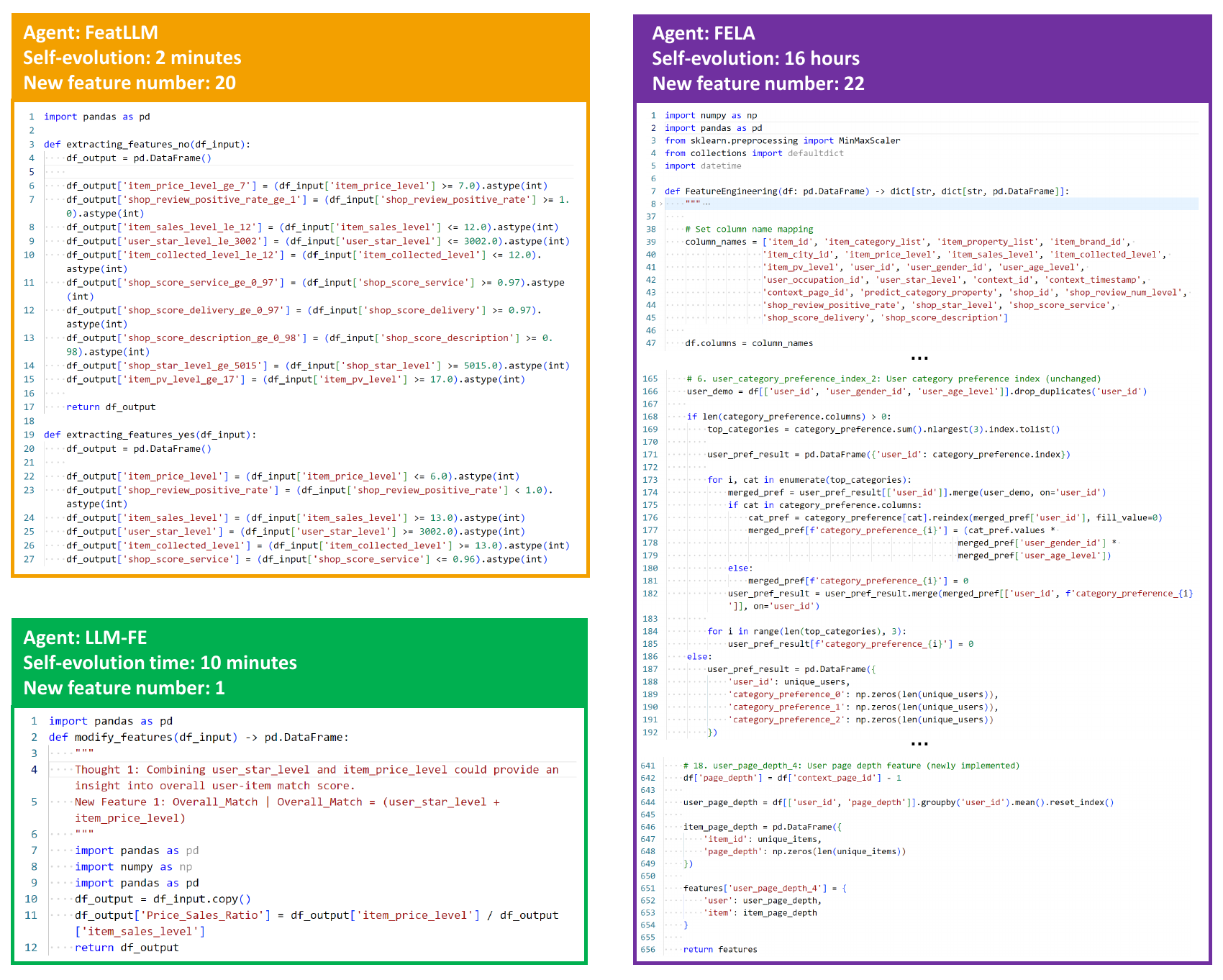}
    \caption{Case study of the generated feature engineering codes}
    \label{fig: case-study code}
\end{figure*}

We begin by analyzing the complexity of the generated feature engineering code on the Taobao dataset, as illustrated in Figure~\ref{fig: case-study code}. {The proposed FELA framework generates 22 features derived from 5 distinct feature ideas, achieving an AUC score of 0.653, an improvement of 0.023 over the raw features. This demonstrates FELA’s capability to efficiently produce a substantial number of high-quality features. In contrast, the FeatLLM baseline generates 20 new features, yet yields negligible AUC improvement compared to raw features. Meanwhile, LLM-FE produces 110 code snippets, with the best-performing code improving AUC by 0.011. However, the optimal LLM-FE code only creates only 1 new feature and lacks a continuous evolution mechanism, ultimately limiting its performance potential. Note that the primary requirement for LLM-agent-based feature engineering is the self-evolution capability with elapsed time \cite{chan2024mle}. The proposed FELA can achieve effective long-horizon self-evolution, which is the key to achieving significant AUC performance improvement. In contrast, the baselines fail to continuously evolve even when increasing the elapsed time.}

\begin{figure*}[!t]
    \centering
    
    \begin{minipage}{0.24\textwidth}  
        \centering
        \includegraphics[width=\linewidth]{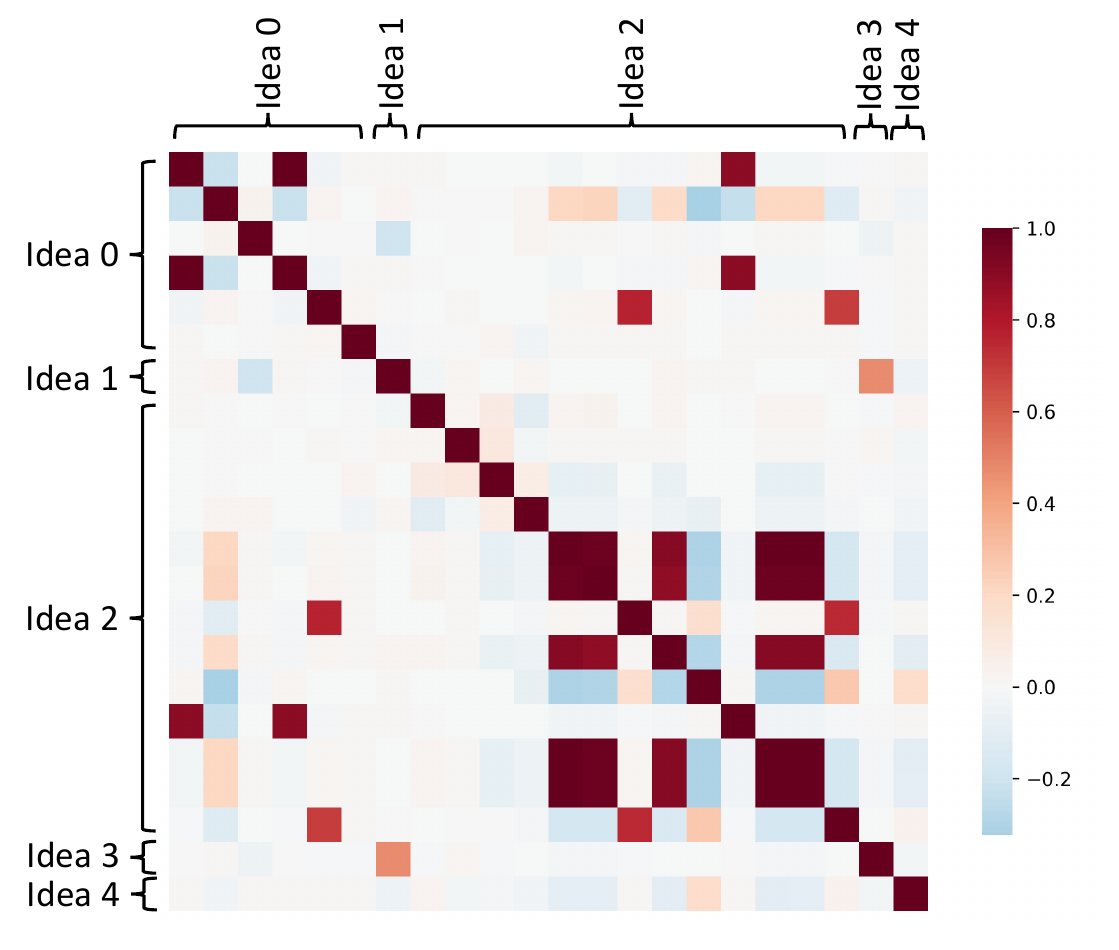}
        \caption{Correlations of selected ideas and features in FELA.}
        \label{fig: case-study correlation}
    \end{minipage}
    \hfill  
    \begin{minipage}{0.24\textwidth}
        \centering
        \includegraphics[width=\linewidth]{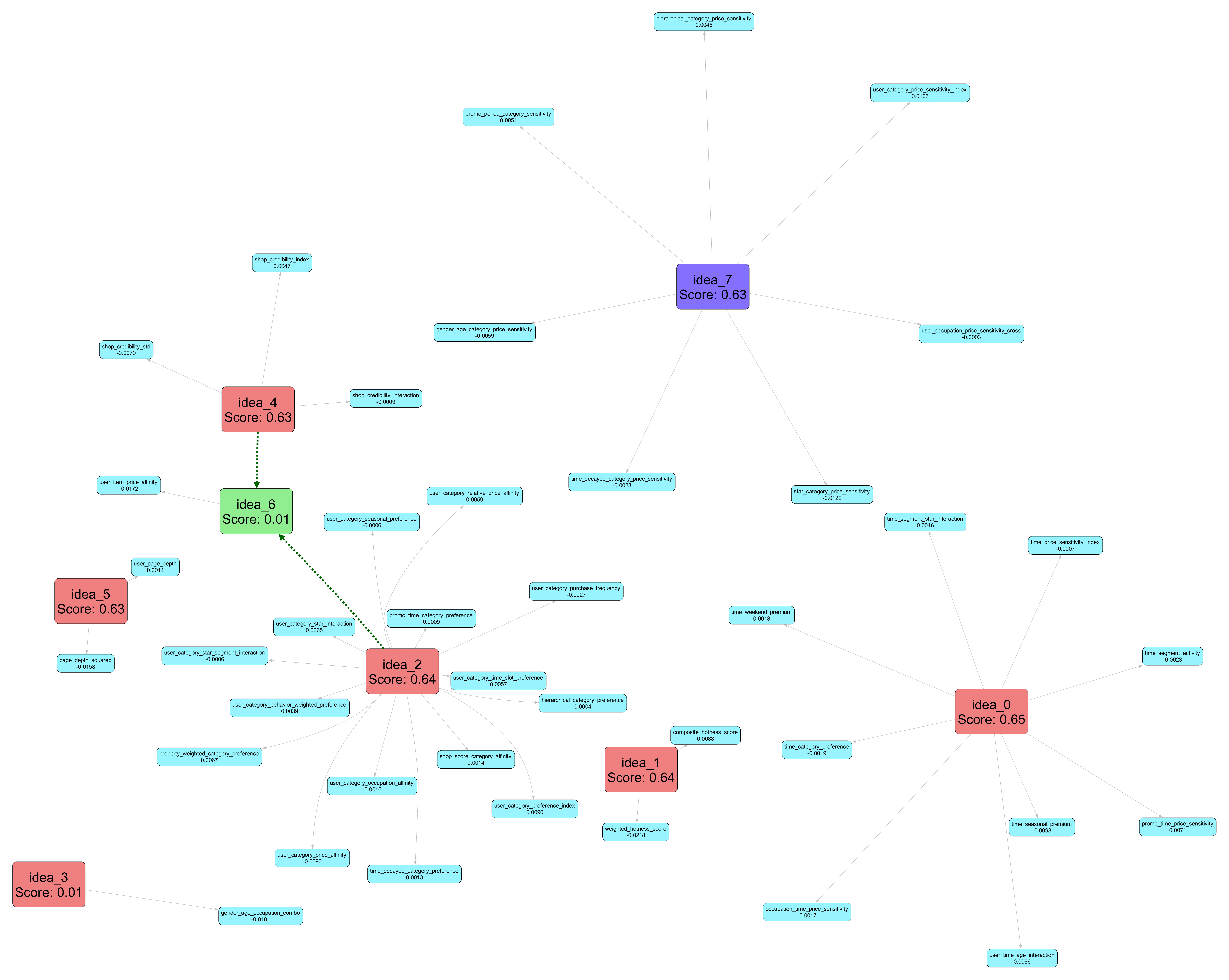}
        \caption{A snapshot of the knowledge database during evolution.}
        \label{fig:big_knowledge_base}
    \end{minipage}
    \hfill  
    \begin{minipage}{0.24\textwidth}
        \centering
        \includegraphics[width=\linewidth]{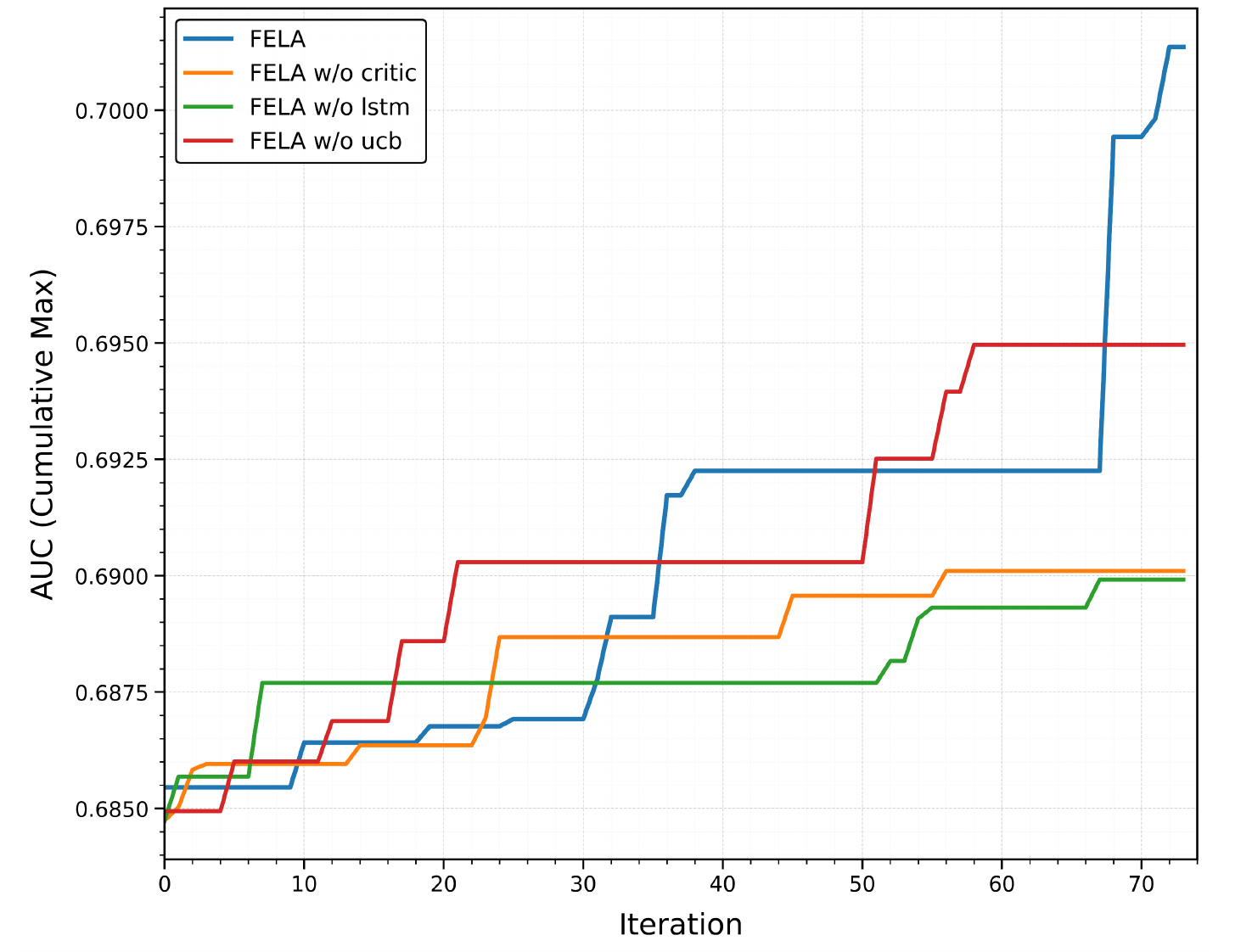}
        \caption{Ablation study of FELA evaluated on the Tencent dataset.}
        \label{fig:ablation}
    \end{minipage}
    \hfill  
    \begin{minipage}{0.24\textwidth}
        \centering
        \includegraphics[width=\linewidth]{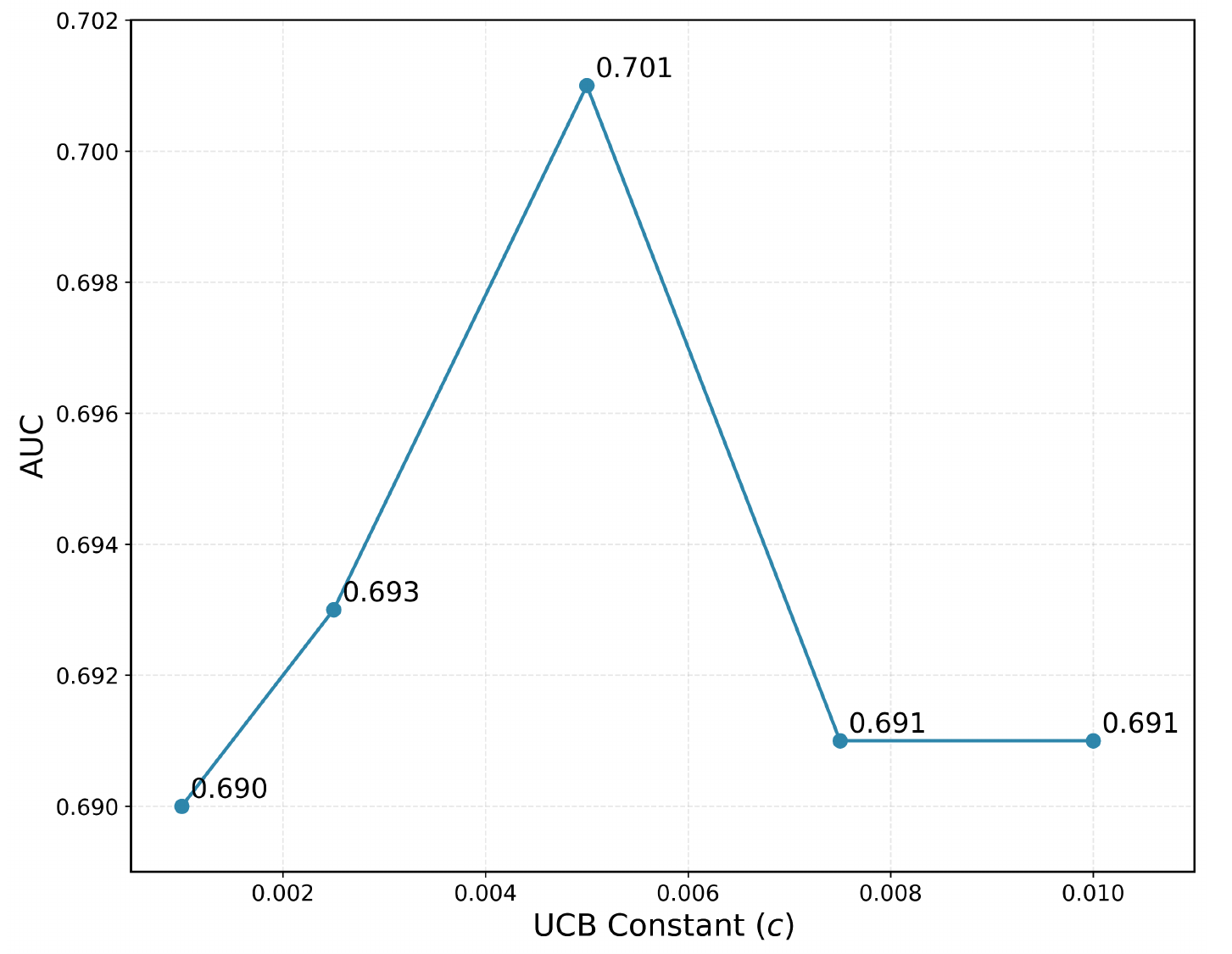}
        \caption{Sensitivity study of UCB-based exploration constant c in FELA}
        \label{fig:sensitivity}
    \end{minipage}
    \vspace{-15pt}
\end{figure*}

We further investigate the feature correlation patterns in FELA by analyzing the Pearson correlation heatmap of the aforementioned 22 features, as visualized in Figure~\ref{fig: case-study correlation}. The analysis reveals that features derived from the same idea exhibit stronger intra-idea correlation. This phenomenon aligns with expectations, as features originating from the same semantic insight naturally share underlying conceptual foundations. Conversely, inter-idea feature correlations remain relatively low, indicating that FELA successfully explores diverse regions of the feature space. This structured exploration strategy demonstrates FELA's capability to conduct steerable feature generation while maintaining semantic interpretability. The framework effectively balances between exploiting productive semantic directions and exploring novel feature combinations, resulting in both discriminative power and explanatory value. In Figure~\ref{fig:big_knowledge_base}, we visualize the sampled knowledge database after evolution on the Taobao dataset. It is a super set of the selected features. Green denotes synthesized ideas. Purple denotes new ideas. Red denotes original ideas. Light blue denotes derived features. Thus, we can easily trace the evolution of the knowledge base structure, as well as understand the evolutionary trajectories between the ideas and associated features.

\subsection{Compatibility to Various Machine Learning Models (RQ3)}

\begin{table}[!t]
\vspace{-10pt}
\caption{{AUC performance comparison with different ML models}} 
\label{tab:ml model}
\centering
\begin{tabular}{ccccc}
\toprule
\textbf{Method/Model} & \textbf{Random Forest} & \textbf{XGBoost} & \textbf{MLP} & \textbf{LR}\\
\midrule
Raw features & 0.683 & 0.648 & 0.666 & 0.650 \\
Skope-rules & 0.685 & 0.647 & 0.556 & 0.653 \\
    OpenFE & 0.686 & 0.655 & 0.501 & 0.650 \\
    CAAFE & 0.686 & 0.650  & 0.558 & 0.650 \\
    AIDE  & 0.683 & 0.648 & 0.666 & 0.650 \\
    FeatLLM & 0.657 & 0.632 & 0.658 & 0.646 \\
    LLM-FE & 0.686 & 0.653 & 0.674 & 0.656 \\
    FELA  & 0.701 & 0.675 & 0.673 & 0.670 \\
\bottomrule
\end{tabular}
\vspace{-20pt}
\end{table}

We investigate the applicability of the proposed FELA over various machine learning models, including Random Forest, XGBoost, MLP, and Logistic classifiers. Then, the AUC performance of the proposed FELA and the baselines over the Tencent dataset is shown in Table~\ref{tab:ml model}. It can be observed that the proposed FELA can achieve the best AUC performance over most of the classifiers, where a 0.012 AUC improvement can be achieved on average. Thus, the proposed FELA system can serve as a robust feature engineering solution for various machine learning models.

\subsection{Abalation Study (RQ4)}
We conduct the ablation study in FELA systems to investigate the contribution of its components, including the critic agent, long-short-term memory, and the UCB-based learning algorithm. Then, the AUC in different iterations for the proposed FELA and its ablations is plotted in Figure~\ref{fig:ablation}. The ablations without the critic agent and the long-short term memory lead to an obvious AUC decline of around 0.01. Thus, the proposed FELA can benefit from long-short-term memory to exploit the evolutionary experience. Additionally, the critic agent can enhance the correctness and consistency of the generated idea and code, which improves the robustness in feature generation and evaluation in FELA. For the ablation without the UCB-based learning algorithm, it can be found that the target AUC monotonically increases at a slower pace and converges to a lower value. Hence, the proposed UCB-based learning can better balance the exploitation and exploration, which enhances the efficiency of the proposed FELA system. 

\subsection{{Sensitivity Study (RQ5)}}
\begin{table}[t]
  \centering
    \belowrulesep=0pt
  \aboverulesep=0pt
  \caption{{Performance of FELA over the Taobao dataset under different LLM models}}
    \begin{tabular}{cccccc}
    \toprule
    LLM model   & Acc   & Prec  & Recall & F1    & AUC \\
    \midrule
    Deepseek-V3 & 0.678 & 0.0294 & 0.551 & 0.0558 & 0.653 \\
    Kimi-K2-Instruct & 0.684 & 0.0287 & 0.526 & 0.0544 & 0.651 \\
    GPT-4o & 0.689 & 0.0294 & 0.530  & 0.0556 & 0.651 \\
    \bottomrule
    \end{tabular}%
  \label{tab:llm backend}%
  \vspace{-10pt}
\end{table}%

{
We begin by evaluating the sensitivity of FELA to different LLM APIs. Employing the APIs of Kimi-K2 \cite{team2025kimi} and GPT-4o \cite{hurst2024gpt}, we report the performance on the Taobao dataset in Table~\ref{tab:llm backend}. The results show that FELA consistently boosts the AUC to over 0.65 across all backends, improving it by more than 0.02 over the raw features. This consistent gain demonstrates that the performance improvement is attributable to FELA's novel architecture, rather than being an artifact of a specific LLM's capabilities.}

{Next, we further consider the sensitivity of FELA over the exploration constant $c$ in the Tencent dataset, which is shown in Fig.~\ref{fig:sensitivity}. It reveals a unimodal relationship between the exploration constant \(c\) and AUC performance, reflecting the classic exploration-exploitation trade-off. Below the optimal value (\(c < 0.005\)), FELA over-exploits operators with high historical scores \(\sum_j s(d_{i,j})/Q_i\), leading to premature convergence. Conversely, when \(c > 0.005\), excessive exploration of under-sampled operators impedes the refinement of promising candidates. The peak at \(c \approx 0.005\) thus represents the ideal balance, maximizing search efficiency and final model performance. Note that proposed FELA can consistently surpass the second optimal baseline Feat-LLM and Open-FE under different $c$, which further validates the system robustness. }

\section{Conclusion \& Future Work}
In this work, we introduced \textbf{FELA (Feature Engineering LLM Agents)}, a comprehensive multi-agent collaboration system designed to automate and enhance feature engineering for complex industrial event log data. FELA integrates large language model agents with an insight-guided self-evolution paradigm, enabling the system to generate novel, explainable, and high-performing features without extensive human intervention. Through the synergistic interaction of \textit{Idea Agents}, \textit{Code Agents}, and \textit{Critic Agents}, the system effectively decomposes the intricate reasoning and implementation processes that traditionally demand significant domain expertise.


Beyond its academic contributions, FELA has also demonstrated substantial real-world impact. The system has been deployed across several internal data-centric applications to enhance business intelligence and operational efficiency. A representative case involves fraud detection in an MMO game with millions of players, where the goal was to improve the accuracy of a scorecard model through richer player-level features. {The baseline includes 60+ expert-crafted features developed by senior data scientists over weeks based on deep domain knowledge. In contrast, we let FELA self-evolve for around 24 hours and keep top 20+ additional features.} By incorporating features derived from FELA, the fraud detection system achieved a \textbf{47.1\% reduction in false positive rate (FPR)} compared to expert-engineered baselines—representing a remarkable improvement with significant commercial value.

{In future research, we will conduct a more rigorous quantitative evaluation of interpretability. First, we will develop a suite of metrics—such as a concept alignment score to measure the stability of feature importance rankings and counterfactual explanation fidelity to assess local linearity—to objectively compare the explainability of FELA-generated features against baseline methods. Second, we will run a controlled human user study with domain experts to quantitatively assess the comprehensibility, usefulness, and actionability of generated features, providing multi-faceted evidence for the superior human-interpretability of our approach. Finally, we plan to explore a hybrid optimization strategy that combines LLM-driven logical structure generation with efficient numerical methods (e.g., coordinate descent) for parameter tuning, further bridging the gap between automated feature engineering and full-scale AutoML and improving both performance and scalability.}

\clearpage
\balance
\bibliographystyle{IEEEtran} 
\bibliography{ref}          

\end{document}